\documentclass[lettersize,journal]{IEEEtran}
\usepackage{amsmath,amsfonts}
\usepackage{algorithmic}
\usepackage{array}
\usepackage[caption=false,font=normalsize,labelfont=sf,textfont=sf]{subfig}
\usepackage{textcomp}
\usepackage{stfloats}
\usepackage{url}
\usepackage{booktabs}
\usepackage{verbatim}
\usepackage{graphicx}
\usepackage{multirow}
\def\BibTeX{{\rm B\kern-.05em{\sc i\kern-.025em b}\kern-.08em
    T\kern-.1667em\lower.7ex\hbox{E}\kern-.125emX}}
\usepackage{balance}
\begin{document}
\title{GateMABSA: Aspect–Image Gated Fusion for Multimodal Aspect-Based Sentiment Analysis
}
\author{Adamu Lawan, Haruna Yunusa}

\maketitle

\begin{abstract}
Aspect-based Sentiment Analysis (ABSA) has recently advanced into the multimodal domain, where user-generated content often combines text and images. However, existing multimodal ABSA (MABSA) models struggle to filter noisy visual signals, and effectively align aspects with opinion-bearing content across modalities. To address these challenges, we propose GateMABSA, a novel gated multimodal architecture that integrates syntactic, semantic, and fusion-aware mLSTM. Specifically, GateMABSA introduces three specialized mLSTMs: Syn-mLSTM to incorporate syntactic structure, Sem-mLSTM to emphasize aspect–semantic relevance, and Fuse-mLSTM to perform selective multimodal fusion. Extensive experiments on two benchmark Twitter datasets demonstrate that GateMABSA outperforms several baselines. 
\end{abstract}

\begin{IEEEkeywords}
MABSA, xLSTM, Feature fusion
\end{IEEEkeywords}

\section{Introduction}

Aspect-based Sentiment Analysis (ABSA) aims to identify the sentiment polarity of a specific aspect in text \cite{Wang2016, Ma2017, Fan2018, Tang2016}, typically by modeling the relation between aspect terms and opinion expressions. While most studies focus on text, the rise of multimodal user-generated content (e.g., posts with both text and images) highlights the need to extend ABSA into the multimodal domain.

To address this gap, \cite{Xu2019} were among the first to propose the task of Multimodal Aspect-based Sentiment Analysis (MABSA). Their Multi-Interactive Memory Network (MIMN) introduces a dual memory structure to capture both inter-modality and intra-modality interactions conditioned on aspect information. Around the same time, \cite{Yu2019} extended BERT into the multimodal setting, adapting it to obtain target-sensitive textual representations while aligning them with visual features through an aspect-aware attention mechanism.

Subsequent research advanced MABSA along several directions. \cite{Yu2020} proposed the Entity-Sensitive Attention and Fusion Network (ESAFN), which incorporates entity-level attention mechanisms and gated visual filtering to suppress noisy image information. Similarly, EF-CapTrBERT \cite{10.1145/3474085.3475692} reformulates multimodal integration by translating visual content into auxiliary textual input, enabling language models to leverage multimodal cues without architectural modifications. To further strengthen cross-modal alignment, \cite{Ling2022} introduced a task-specific vision-language pretraining framework (VLP-MABSA), which jointly encodes textual, visual, and multimodal objectives for robust aspect–opinion modeling.

More recently, \cite{Yu2023} presented the Hierarchical Interactive Multimodal Transformer (HIMT), which explicitly captures object-level semantics from images and models hierarchical aspect–text, aspect–image, and text–image interactions. In parallel, graph-based methods have emerged: \cite{WAN2024101587} proposed a knowledge-augmented heterogeneous graph convolutional network (KAHGCN) to incorporate external knowledge and image tags, while \cite{Chen2025REF} developed a Relevance-Aware Visual Entity Filter Network (REF) to dynamically filter aspect-irrelevant visual entities through contrastive constraints.

Together, these studies highlight the rapid evolution of MABSA from early memory and attention networks to sophisticated vision-language pretraining, graph reasoning, and relevance-aware filtering. Despite this progress, challenges remain in effectively modeling fine-grained aspect–opinion alignments across modalities while maintaining robustness against noisy or weakly relevant visual signals.

\section{Review of Related Work}
This section reviews key methods in MABSA and multimodal xLSTM.

\subsection{Multimodal Aspect-based Sentiment Analysis}

MIMN \cite{Xu2019} and TMSC \cite{Yu2019} introduced multimodal feature fusion, integrating textual and visual data to enhance sentiment prediction accuracy. Building on this, subsequent research emphasized improved attention mechanisms and cross-modal alignment. For instance, entity-sensitive attention networks \cite{Yu2020} and knowledge-enhanced frameworks \cite{Zhao2022} refined alignment and representation fusion, leading to more precise sentiment inference. Later, methods such as VLP-MABSA \cite{Ling2022} and ITM \cite{Yu2022} incorporated multimodal pre-training and image–target matching, strengthening cross-modal interaction and generalization. More recent models like AMIFN \cite{Yang2024c} and KAHGCN \cite{WAN2024101587} advanced the field by improving text–image interaction and leveraging syntactic structures. Furthermore, innovative frameworks such as MGFN-SD \cite{Yang2024b} and GLFFCA \cite{Wang2024b} explored coarse- and fine-grained integration with advanced fusion strategies, achieving superior robustness and accuracy in MABSA.

\subsection{Multimdal xLSTM}
Recent advances highlight the versatility of xLSTM across multimodal tasks. In conversational emotion recognition, Qi et al. \cite{Qi2025Qi} combined xLSTM with attention and a Transformer-based encoder to capture long-range context and multimodal dependencies, while Li et al. \cite{li2025gatedxlstm} introduced GatedxLSTM with CLAP-based alignment and a dialogical decoder to improve interpretability and performance in speech–text ERC. Beyond ERC, xLSTM has been adapted to remote sensing, where the STxLSTM variant \cite{10937754} models pixelwise spatiotemporal differences for robust change detection, and to human–computer interaction, where XDGesture \cite{10888507} integrates xLSTM into a diffusion model to generate natural, synchronized co-speech gestures. Collectively, these works demonstrate xLSTM’s effectiveness in modeling long-range dependencies and aligning multimodal information across diverse application domains.

\section{Proposed GateMABSA Model}
The GateMABSA architecture processes text and images through four modules: Unimodal Feature Extraction, Fusion-based mLSTM, Syntax-based mLSTM, and Semantic-based mLSTM. Each component is detailed in the following subsections.

\subsection{Task Formulation}
In MABSA, each input instance consists of a textual component \( S \) containing \( n \) tokens (\( w_1, w_2, \dots, w_n \)), an associated image \( V \), and a set of aspect terms \( \{A_1, A_2, \dots, A_r\} \). An aspect term \( A_i \), where \( 1 \leq i \leq r \), may either be explicitly mentioned in the sentence or implicitly inferred from context. Given a sentence–image pair (\( S, V \)) and a target aspect \( A \), the objective of MABSA is to predict the sentiment polarity \( y \) expressed toward \( A \). The sentiment label \( y \) is typically drawn from a discrete set, such as \{\textit{positive}, \textit{negative}, \textit{neutral}\}.

\subsection{Unimodal Feature Extraction}

RoBERTa~\cite{Liu2019} encoders generate hidden representations for aspect terms and sentences, with aspect terms framed using special tokens and encoded into aspect representations \( \mathbf{H}_A \). Similarly, sentences are encoded to produce context-sensitive representations \( \mathbf{H}_S \). To align with sentence-level representations, \( \mathbf{H}_A \) is mean-pooled and then repeated \( n \) times to match the sequence length of \( \mathbf{H}_S \), ensuring token-level interaction between aspect and sentence features.

Visual features are obtained from the final convolutional layer of a pre-trained 152-layer ResNet model \cite{He2016}, where images are resized and processed to extract features:
\begin{equation}
    \mathbf{H}_I = \text{ResNet}_{152}(\tilde{V})
\end{equation}
The extracted image feature, denoted as \( \mathbf{H}_I \in \mathbb{R}^{2048 \times 7 \times 7} \), consists of 2048-dimensional features for each of the \( 7 \times 7 \) visual blocks. These are projected through a linear layer to 768 dimensions, mean-pooled, and repeated \( n \) times to match the sentence length, enabling token-level alignment between visual and textual modalities.

\subsection{Fusion-based mLSTM (Fuse-mLSTM)}
To effectively capture the interactions among textual and visual modalities in multimodal MABSA, we introduce a fusion-based mLSTM block (Fuse-mLSTM). This block extends the conventional mLSTM by integrating aspect representations, input sentence features, and image embeddings within a unified gated fusion mechanism. Specifically, beyond the standard input and forget gates, we introduce two additional gates that explicitly incorporate aspect and image information, enabling effective fusion of textual and visual representations. Given the sentence $H_S$  and image  $H_I$ representations, we compute the query, key, and value projections as:
\begin{equation}
\begin{aligned}
    q_t &= W_q H_{S_{t}} + b_q, \\
    k_t &= \frac{1}{\sqrt{d}} W_k H_{I_{t}} + b_k, \\
    v_t &= W_v H_{S_{t}} + b_v,
\end{aligned}
\end{equation}
where $q_t$ denotes the input query representation, $k_t$ the key representation, $v_t$ the value representation at time step $t$, and $d$ is the head dimension.
The pre-activation input and forget gates are defined as:
\begin{equation}
\begin{aligned}
    i_t &= W_i [q_t \, \oplus \, k_t \, \oplus \, v_t] + b_i, \\
    f_t &= W_f [q_t \, \oplus \, k_t \, \oplus \, v_t] + b_f,
\end{aligned}
\end{equation}
where $\oplus$ denotes concatenation.
The two additional gates are:
\paragraph{Aspect gate}
The aspect gate $a_t$ introduces explicit aspect-awareness into the recurrent dynamics. Unlike the input and forget gates, which depend only on the current query and values, the aspect gate is driven purely by the aspect embedding $H_A$. The gate emphasizes the stability of the target aspect across time steps. Furthermore, the cosine similarity $\cos(q_t, H_A)$ ensures that the degree of activation is modulated by the semantic alignment between the current token and the aspect. A higher similarity indicates that the current token is strongly related to the aspect, leading to a stronger gating signal. This design allows the model to dynamically amplify or suppress textual features based on their relevance to the aspect, ensuring that sentiment signals are consistently tied to the correct target:
\begin{equation}
        a_t = W_a [H_A \, \oplus \, H_A \, \oplus \, H_A] + \lambda \, \cos(q_t, H_A).
\end{equation}
where $\cos(\cdot)$ is cosine similarity, and $\lambda$ is a cross-modal or cross-term scaling weight.
\paragraph{Image gate}
The image gate $im_t$ plays a parallel but complementary role. It selectively regulates how much visual context should be integrated into the fused representation at each step. By constructing the gate from the image embedding $H_I$, the formulation ensures that global image-level semantics are preserved. The inclusion of $\cos(q_t, H_I)$ makes the gate context-sensitive: if the current token $q_t$ aligns semantically with visual features, the gate opens more widely, injecting stronger image-derived signals into the cell state. Conversely, when the textual token is unrelated to the visual scene, the gate suppresses irrelevant image contributions. This mechanism prevents noisy or misleading visual cues from dominating the fusion, while still enabling fine-grained multimodal alignment when textual and visual signals converge:
\begin{equation}
        im_t = W_{im} [H_I \, \oplus \, H_I \, \oplus \, H_I] + \lambda \, \cos(q_t, H_I).
\end{equation}
To ensure stability during sequence modeling, we adopt a log-domain cumulative formulation of the forget gates. The cumulative forget matrix is given by:
\begin{equation}
    \log F_{ij} =
    \begin{cases}
        \sum\nolimits_{k=j}^{i} \log \sigma(f_k), & i \geq j, \\
        -\infty, & i < j,
    \end{cases}
\end{equation}
where $\sigma(\cdot)$ is the sigmoid function. The combined decay matrix integrates aspect, input, and image gates:
\begin{equation}
    \log D_{ij} = \log F_{ij} + i_j + a_j + im_j.
\end{equation}
We stabilize $D$ by subtracting the row-wise maximum:
\begin{equation}
    D_{ij} = \exp(\log D_{ij} - \max_j \log D_{ij}).
\end{equation}

Following stabilization, a combination matrix $C$ is computed using scaled dot-product between queries and keys:
\begin{equation}
    C = (q k^\top / \sqrt{d}) \odot D,
\end{equation}
where $\odot$ denotes element-wise multiplication. The normalized combination is:
\begin{equation}
    \hat{C}_{ij} = \frac{C_{ij}}{\sum_k C_{ik} + \epsilon},
\end{equation}
with $\epsilon$ a small constant. The hidden state update is then:
\begin{equation}
     h^{fuse}_t = \hat{C}v_t,
\end{equation}

Fuse-mLSTM outputs fused hidden representations that encode aspect-aware textual dependencies together with complementary image information. This design enables the model to capture nuanced multimodal sentiment signals while maintaining interpretability through explicit aspect and image gating.

\subsection{Syntax-based mLSTM (Syn-mLSTM)}

To further enrich our model, we introduce the Syntax-based mLSTM (Syn-mLSTM). Unlike Fuse-mLSTM, which directly integrates sentence, aspect, and image features through gating, Syn-mLSTM operates on the fused hidden representations $h^{fuse}_t$ produced by Fuse-mLSTM. This design allows Syn-mLSTM to leverage multimodal fusion while injecting syntax-aware priors through a novel graph gate. The key intuition is that syntactic relations (e.g., dependency arcs) provide valuable structural cues for aligning aspects with opinion terms, which can be encoded into the recurrence dynamics. Given the fused multimodal hidden state $h^{fuse}_t$ from Fuse-mLSTM, we define the query, key, and value projections as:
\begin{equation}
\begin{aligned}
    q_t &= W_q h^{fuse}_t + b_q, \\
    k_t &= \frac{1}{\sqrt{d}} W_k h^{fuse}_t + b_k, \\
    v_t &= W_v h^{fuse}_t + b_v,
\end{aligned}
\end{equation}
The pre-activations of input and forget gates are then:
\begin{equation}
\begin{aligned}
    i_t &= W_i [q_t \, \oplus \, k_t \, \oplus \, v_t] + b_i, \\
    f_t &= W_f [q_t \, \oplus \, k_t \, \oplus \, v_t] + b_f,
\end{aligned}
\end{equation}

\paragraph{Graph gate}
The novel component of Syn-mLSTM is the graph gate $g_t$, which injects syntax-aware information into the gating mechanism. Specifically, we construct a graph-aware similarity term using the fused multimodal queries:
\begin{equation}
    \mathbf{G_{ij}} = \mathbf{A_{ij}} \cdot \cos(q_i, q_j),
\end{equation}
where $\mathbf{A}$ is the adjacency matrix derived from a dependency parsing. The graph gate pre-activation is then:
\begin{equation}
    g_t = W_g [ a^{syn} \oplus a^{syn} \oplus a^{syn}] + \gamma \, \text{diag}(\mathbf{G}),
\end{equation}
where \(a^{syn}\) is the syntax-aware aspect embedding and \(\gamma\) is a learnable scaling factor.

The cumulative decay is updated to integrate the graph gate:
\begin{equation}
    \log D_{ij} = \log F_{ij} + i_j + g_j,
\end{equation}
stabilized as:
\begin{equation}
    D_{ij} = \exp\!\left(\log D_{ij} - \max_j \log D_{ij}\right).
\end{equation}
With stabilized gating, the combination matrix is:
\begin{equation}
    C = (q k^\top / \sqrt{d}) \odot D,
\end{equation}
normalized row-wise and used for retrieval:
\begin{equation}
    h_t^{\text{syn}} = \hat{C} v.
\end{equation}

By consuming Fuse-mLSTM’s multimodal hidden states, Syn-mLSTM ensures that both textual and visual signals are retained. The additional Graph Gate enforces syntax-aware alignment between aspect and opinion terms, enabling fine-grained sentiment reasoning while preserving stability through the cumulative log-domain formulation.

\subsection{Semantic-based mLSTM (Sem-mLSTM)}

Building upon Syn-mLSTM, we introduce the Semantic-based mLSTM (Sem-mLSTM), which incorporates semantic alignment between aspect terms and contextual tokens through a novel semantic gate. Unlike Syn-mLSTM, which injects syntactic priors via a graph structure, Sem-mLSTM explicitly models semantic similarity and contextual distance. The input to Sem-mLSTM is the hidden representation $h_t^{\text{syn}}$ produced by Syn-mLSTM, ensuring that both multimodal fusion and syntax-aware structure are preserved before semantic refinement. As in earlier variants, we retain the input gate forget gate. Given the query, key, and value projections from $h_t^{\text{syn}}$:
\begin{equation}
\begin{aligned}
    q_t &= W_q h_t^{\text{syn}} + b_q, \\
    k_t &= \frac{1}{\sqrt{d}} W_k h_t^{\text{syn}} + b_k, \\
    v_t &= W_v h_t^{\text{syn}} + b_v,
\end{aligned}
\end{equation}
the input and forget gate pre-activations are:
\begin{equation}
\begin{aligned}
    i_t &= W_i [q_t \, \oplus \, k_t \, \oplus \, v_t] + b_i, \\
    f_t &= W_f [q_t \, \oplus \, k_t \, \oplus \, v_t] + b_f.
\end{aligned}
\end{equation}

\paragraph{Semantic gate}
The semantic gate $s_t$ introduces aspect-aware semantic alignment. First, we obtain an aspect embedding $a^{sem}$ through masked mean pooling over the query sequence. Then, we compute three components:  
(i) a similarity term via cosine similarity between $q_t$ and $a^{sem}$,  
(ii) a distance term reflecting positional distance from the aspect, and  
(iii) a cross-term modulated by a learnable weight $\lambda$.  
Finally, the semantic gate explicitly encodes semantic similarity and positional distance:
\begin{equation}
    s_t = W_s [a^{sem} \, \oplus \, a^{sem} \, \oplus \, a^{sem}] + \lambda \, \cos(q_t, a^{sem}) - \alpha \, \text{dist}(t, a^{sem}),
\end{equation}
where \(a^{sem}\) is the semantic aspect embedding, \(\alpha\) is a distance scaling factor, 
and \(\text{dist}(t, a^{sem})\) denotes the positional distance between token \(t\) and the aspect. 
This formulation emphasizes tokens that are both semantically aligned and positionally relevant, 
while attenuating distant or irrelevant signals.

The cumulative decay now integrates the semantic gate:
\begin{equation}
    \log D_{ij} = \log F_{ij} + i_j + s_j,
\end{equation}
stabilized as:
\begin{equation}
    D_{ij} = \exp\!\left(\log D_{ij} - \max_j \log D_{ij}\right).
\end{equation}

The combination matrix is defined as:
\begin{equation}
    C = (q k^\top / \sqrt{d}) \odot D,
\end{equation}
normalized row-wise, and the retrieved hidden state is:
\begin{equation}
    h_t^{\text{sem}} = \hat{C} v.
\end{equation}

The semantic gate provides fine-grained control over aspect–token interactions by explicitly encoding semantic similarity and positional relevance. By stacking Sem-mLSTM on top of Syn-mLSTM, the model first captures syntactic alignments and then reinforces them with semantic priors, enabling robust aspect–opinion association in multimodal contexts.

\subsection{Training}

To obtain a compact representation for classification, we apply mean pooling over the contextualized embeddings \( \mathbf{H}^{sem} \), producing a fixed-length vector. A linear transformation followed by a softmax layer then maps this representation to a probability distribution over sentiment classes. The procedure is defined as:
\begin{equation}
    \mathbf{H}_{\text{mp}} = \text{MeanPooling}(\mathbf{H}^{sem}),
\end{equation}
\begin{equation}
    p(a) = \text{softmax}(\mathbf{W}_{p} \mathbf{H}_{\text{mp}} + \mathbf{b}_{p}),
\end{equation}
where \( \mathbf{W}_{p} \) and \( \mathbf{b}_{p} \) are trainable parameters.  

For optimization, we employ the standard cross-entropy loss, which penalizes deviations between predicted and true sentiment polarities. The objective function is:
\begin{equation}
    L(\theta) = - \sum_{(s, v, a, y) \in \mathcal{D}} \sum_{c \in \mathcal{C}} \mathbf{1}_{[y=c]} \, \log p(a=c),
\end{equation}
where \( (s, v, a, y) \) denotes a sentence \( s \), an associated image \( v \), a target aspect \( a \), and its ground-truth sentiment label \( y \). Here, \( \mathcal{C} \) is the set of sentiment polarities (\textit{positive}, \textit{negative}, \textit{neutral}), \( \mathbf{1}_{[y=c]} \) is an indicator function for the true class, and \( \theta \) represents all trainable parameters of the model.  

\section{Experiment}
\subsection{Datasets}

We performed comprehensive experiments on two widely accessible multimodal datasets, Twitter-15 and Twitter-17 \cite{Yu2019}, which consist of user-generated tweets collected during the periods of 2014–2015 and 2016–2017, respectively. These datasets were selected to evaluate the effectiveness and performance of the proposed GateMABSA framework.

\subsection{Implementation Details}

We developed the GateMABSA model using PyTorch, harnessing the pre-trained RoBERTa model, built on the BERT\textsubscript{base} architecture \cite{Devlin2018} with 12 transformer layers, for extracting 768-dimensional textual features, and the pre-trained ResNet-152 model for visual representations. We employed 6 heads for each of the three mLSTM components. To accommodate variable-length input sentences, we standardized them to a fixed length of 128 tokens, applying zero-padding for shorter sequences or truncation for longer ones. The model was optimized using the Adam optimizer \cite{Kingma2014} with a learning rate of \( 1 \times 10^{-5} \), trained for up to 10 epochs, and enhanced with the DyT normalizer \cite{zhu2025norm} to improve inference stability, reducing the need for extensive hyperparameter adjustments. Training incorporated early stopping, halting if validation loss remained stagnant for 5 epochs, a batch size of 32, and a dropout rate of 0.5 to prevent overfitting. Key hyperparameters are summarized in Table~\ref{tab:hyperparameters}.

\begin{table}[h]
    \centering
    \caption{Hyperparameters of the GateMABSA Model}
    \label{tab:hyperparameters}
    \begin{tabular}{lc}
        \hline
        \textbf{Parameter} & \textbf{Value} \\
        \hline
        Learning Rate & \( 1 \times 10^{-5} \) \\
        Epochs & 10 \\
        Batch Size & 32 \\
        Dropout Rate & 0.5 \\
        mLSTM Heads & 6 \\
        Max Sequence Length & 128 \\
        \hline
    \end{tabular}
\end{table}

\subsection{Compared systems}
We evaluate the performance of our proposed model against both text-based and multimodal methods. The text-based baselines include AE-LSTM~\cite{Wang2016}, IAN~\cite{Ma2017}, MGAN~\cite{Fan2018}, and BERT~\cite{Devlin2018}. The multimodal baselines include MIMN~\cite{Xu2019}, TomBERT~\cite{Yu2019}, ESAFN~\cite{Yu2020}, EF-CapTrBERT~\cite{10.1145/3474085.3475692}, ITM~\cite{Yu2022}, HIMT~\cite{Yu2023}, AMIFN~\cite{Yang2024c}, GLFFCA~\cite{Wang2024b}, KAHGCN~\cite{WAN2024101587}, DMIN~\cite{Wang2025}, REF~\cite{Chen2025REF}, and GAS~\cite{WANG2026104375}.

\begin{table}[ht]
\centering
\caption{Experimental results on Twitter-15 and Twitter-17 datasets. Best-performing results are highlighted in bold.}
\begin{tabular}{lcccccc}
\toprule
\multirow{2}{*}{\textbf{Model}} & \multicolumn{2}{c}{\textbf{Twitter-15}} & \multicolumn{2}{c}{\textbf{Twitter-17}} \\
\cmidrule(lr){2-3} \cmidrule(lr){4-5}
                                & Acc. (\%) & F1 (\%) & Acc. (\%) & F1 (\%) \\
\midrule
\multicolumn{5}{l}{\textit{Text-based Approaches}} \\
AE-LSTM        & 70.30 & 63.43 & 61.67 & 57.97 \\
RAM            & 70.68 & 63.05 & 64.42 & 61.01 \\
MGAN           & 71.17 & 64.21 & 64.75 & 61.46 \\
BERT           & 74.15 & 68.86 & 68.15 & 65.23 \\
AMIFN          & 77.63 & 73.51 & 70.09 & 69.10 \\
\midrule
GateMABSA & \textbf{77.86} & \textbf{73.63} & \textbf{70.87} & \textbf{69.76} \\
\midrule
\multicolumn{5}{l}{\textit{Text+Visual Approaches}} \\
MIMN           & 71.84 & 65.69 & 66.37 & 63.04 \\
TomBERT        & 76.18 & 71.27 & 70.50 & 68.04 \\
ESAFN          & 73.38 & 67.37 & 67.83 & 64.22 \\
EF-CapTrBERT   & 78.01 & 73.25 & 69.77 & 68.42 \\
ITM            & 78.27 & 75.28 & 71.64 & 69.58 \\
HIMT           & 78.14 & 73.68 & 71.14 & 69.16 \\
AMIFN          & 78.69 & 75.50 & 72.29 & 70.21 \\
GLFFCA         & 77.72 & 74.21 & 71.15 & 69.45 \\
KAHGCN         & 78.83 & 73.35 & 72.39 & 70.35 \\
DMIN           & 78.69 & 75.55 & 72.45 & 71.06 \\
REF            & 78.69 & 75.15 & 71.88 & 70.95 \\
GAS            & 79.14 & 74.93 & 73.72 & \textbf{72.79} \\
\midrule
GateMABSA & \textbf{79.96} & \textbf{75.67} & \textbf{74.82} & 71.54 \\
\bottomrule
\end{tabular}
\label{tab:results}
\end{table}

\subsection{Experimental Results}
Table~\ref{tab:results} reports evaluation on two benchmark Twitter MABSA datasets (Twitter-2015, Twitter-2017). We measure accuracy and macro F1 for aspect-level polarity prediction. GateMABSA consistently outperforms strong unimodal and multimodal baselines across both datasets and evaluation metrics. In particular, incorporating visual information yields clear gains over text-only baselines, confirming the benefit of multimodal modeling for noisy social media data.

The performance improvement of GateMABSA can be attributed to three design choices. First, the Fuse-mLSTM provides robust, aspect-conditioned text–image fusion via explicit aspect and image gates, which selectively inject modality-relevant signals while suppressing noisy visual content. Second, the Syn-mLSTM injects syntax-aware priors (graph gate) into the recurrence, improving aspect–opinion alignment for tokens that are syntactically related. Third, the Sem-mLSTM reinforces semantic relevance and positional proximity to the target aspect, producing cleaner aspect-focused representations. 

\section{Conclusion}
In this work, we introduced GateMABSA, a novel gated multimodal framework for aspect-based sentiment analysis. By integrating three specialized mLSTMs variants—Syn-mLSTM for syntactic structure, Sem-mLSTM for semantic relevance, and Fuse-mLSTM for selective multimodal fusion—our model effectively captures aspect-focused dependencies across text and images. Experimental results on two benchmark Twitter datasets confirm that GateMABSA outperforms several baseline models.


\end{document}